\documentclass[10pt, a4paper]{article}

\usepackage[final]{lrec2026} 
\usepackage{inconsolata}
\usepackage{amsmath}
\usepackage{tablefootnote}
\usepackage{gb4e}
\noautomath

\usepackage{booktabs}
\usepackage{multirow} 
\usepackage{soul}
\usepackage{pifont}
\usepackage{makecell,array}
\newcommand{\res}[2]{\makecell[l]{#1\\ {\citeplanguageresource{#2}}}}

\definecolor{copexp}{rgb}{1.0, 0.8, 0.8}
\definecolor{synonyms}{rgb}{0.8, 1.0, 0.8}
\definecolor{tensem}{rgb}{0.8, 0.8, 1.0}
\definecolor{wft}{rgb}{1.0, 1.0, 0.8}
\definecolor{translit}{rgb}{1.0, 0.8, 1.0}

\newcommand{\error}[2]{\sethlcolor{#1}\hl{#2}}
\usepackage{fontawesome5}

\title{RILEC: Detection and Generation of L1 Russian Interference Errors in English Learner Texts}

\name{Darya Kharlamova\textsuperscript{\faCrow} and Irina Proskurina\textsuperscript{\faLandmark} }
\address{\textsuperscript{\faCrow}Higher School of Economics \\
\textsuperscript{\faLandmark}Université Claude Bernard Lyon 1, Université Lumière Lyon 2, ERIC\\
        dasha.kh18@gmail.com\\}

\abstract{
Many errors in student essays can be explained by influence from the native language (L1). L1 interference refers to errors influenced by a speaker’s first language, such as using \textit{stadion} instead of \textit{stadium}, reflecting lexical transliteration from Russian.
In this work, we address the task of detecting such errors in English essays written by Russian-speaking learners. We introduce \textsc{RILEC}, a large-scale dataset of over 18,000 sentences, combining expert-annotated data from \textsc{REALEC} with synthetic examples generated through rule-based and neural augmentation. We propose a framework for generating L1-motivated errors using generative language models optimized with PPO, prompt-based control, and rule-based patterns. Models fine-tuned on \textsc{RILEC} achieve strong performance, particularly on word-level interference types such as transliteration and tense semantics. We find that the proposed augmentation pipeline leads to a significant performance improvement, making it a potentially valuable tool for learners and teachers to more effectively identify and address such errors.
 \\ \newline  }

\begin{document}

\maketitleabstract

\section{Introduction}

The influence of our native language (L1) on writing in a second language is manifold. 
It can lead to pronunciation-induced spelling errors, lexical choices based on the mother tongue, and tense or word order decisions that mirror L1 grammar~\cite{odlin2003cross}. 

\definecolor{copexp}{rgb}{1.0, 0.8, 0.8}
\definecolor{synonyms}{rgb}{0.8, 1.0, 0.8}
\definecolor{tensem}{rgb}{0.8, 0.8, 1.0}
\definecolor{wft}{rgb}{1.0, 1.0, 0.8}
\definecolor{translit}{rgb}{1.0, 0.8, 1.0}

Some examples of native language interference in second-language acquisition include the absence or unconventional usage of certain grammatical categories, such as articles for Russian speakers and auxiliary verb constructions for German speakers, misused structures, such as confusion between \textit{have} and \textit{be} among French and Spanish speakers, and the incorrect use of \textit{will} in conditionals, particularly by Russian and Spanish learners.

Features extracted from a learner’s first language (L1) have been shown to enhance error detection and correction tools designed for language learners \cite{chang2008automatic, ng2014conll}. For instance, native language information has been shown to influence the performance of grammatical error classifiers targeting errors commonly made by non-native English learners \cite{rozovskaya2017adapting}. It is also known to improve Large Language Model (LLM)-powered tools designed to support reading comprehension and writing skills \cite{antoniou2024using,heck2023relevance}.
\begin{table}[h]
\centering
{\footnotesize
\begin{tabular}{p{0.45\textwidth}}
\toprule
\midrule
We will pay more attention to our health if we \error{tensem}{will have} enough time.  The straightforward approach is for people to pay attention \error{synonyms}{on} their nutrition and daily activity, and to refuse \error{synonyms}{from} using alcohol and tobacco.  
On one hand, \error{copexp}{most of} parents want to shield the younger generation from \error{synonyms}{another} cureless diseases, including ads between TV shows or \error{wft}{youtube’s} videos.  
On the other hand, we have \error{translit}{firm} and organizations producing unhealthy goods, and these \error{translit}{firms} want to expand their base of consumers through advertising.
\\
\bottomrule
\end{tabular}
}
\caption{An excerpt from a real essay parsed by the model fine-tuned on \textsc{RILEC}: \error{synonyms}{Synonyms}, \error{copexp}{Copying Expression}, \error{tensem}{Tense Semantics}, \error{wft}{Word Form Transmission}, and \error{translit}{Transliteration}.}

\label{tab:ungrammatical_examples_intro}
\end{table}

Although much research has been done on grammatical error detection (GED), simply identifying ungrammatical structures does not reveal the underlying causes of these errors. As a result,
existing tools which detect and correct errors but do not explain their causes hinder the learning process more than help it \cite{brown2014principles}. Information about the possible reasons behind the errors is also important for teachers to address specific difficulties, adjust study plans, create helpful learning materials, and score essays more efficiently.
In addition, identifying L1-motivated errors can be beneficial for students at any proficiency level, as lower-level students might not recognize these errors independently, and even trained professionals sometimes struggle to detect them reliably.

In this paper, we investigate the problem of classifying potential causes of L1 interference.
We use a linguistic interference tagging system shown in \autoref{tab:ungrammatical_examples_intro} to annotate our dataset, drawing on the research into interference done by \citealp{weinreich1979languages}.

\begin{table*}[ht!]
\centering
\scriptsize
\begin{tabular}{l p{3cm} c c c c}
\toprule
\textbf{Dataset} & \textbf{L1 Background} & \textbf{Sents.} & \textbf{Error Tags} & \textbf{CEFR Level} & \textbf{Task} \\
\midrule
\res{FCE}{yannakoudakis2018developing} & NA, Mixed L1s & 2,695 & 71 & B1-B2 & GEC/GED \\
\res{KJ}{nagata2011creating} & JA & 3,199 & 22 & A1-A2? & GEC/GED \\
\res{CoNLL-2014}{ng-etal-2014-conll} & NA, Mixed L1s & 1,312 & 28 & C1 & GEC/GED \\
\res{JFLEG}{napoles-etal-2017-jfleg} & NA, Mixed L1s & 747 & - & A1-C2? & GEC\\
\res{BEA-2019}{bryant-etal-2019-bea} & NA, Mixed L1s (NUS students; EN, ZH, MS, TA) & 4,384 & 25 & A1-Native & GEC/GED \\
\res{ICNALE}{ishikawa2023icnale} & 10 Asian L1s (ZH, ID, JA, KO, TH, ZH-TW, HK-YUE, PK-UR, PH-TL, TA) & 15,000 essays & - & A2-C1 & GEC\\
\res{ICLEv3}{granger2020iclev3} & 26 L1 (BG, ZH, CS, NL, FI, FR, DE, IT, JA, NO, PL, RU, ES, SV, TR, TN, ...) & 9,529 essays & - & B2-C2 & GEC/NLI \\
\res{TOEFL11}{blanchard2013toefl11} & 11 L1s (AR, ZH, FR, DE, HI, IT, JA, KO, ES, TE, TR) & 12,100 essays & - & A2-C1 & GEC/NLI \\
\res{REALEC}{vinogradova2022review} & RU & 18,710 essays & $\sim$50 & B1-C1 & GEC/GED \\
\midrule
\textbf{RILEC} & \textbf{RU} & \textbf{18,830} & \textbf{5} & \textbf{B1-B2} & \textbf{L1 GED} \\
\bottomrule
\end{tabular}
\caption{Comparison of English learner and GEC corpora by L1 background, dataset size, tag inventory, CEFR range, and task. Only RILEC includes explicit L1-specific interference tags. 
GEC = Grammatical Error Correction. 
GED = Grammatical Error Detection. 
NLI = Native Language Identification. 
L1 GED = L1-interference Error Detection. 
A question mark indicates unknown or approximate information.}
\label{tab:rilec_corpora_comparison}
\end{table*}
Our main contributions are the following:  
(1)~We introduce \textbf{\textsc{RILEC}} (\textbf{R}ussian L1 \textbf{I}nterference \textbf{L}earner \textbf{E}nglish \textbf{C}orpus), the first large-scale dataset of L1-motivated errors, containing over 18,000 sentences, both real and synthetically augmented. \footnote{\url{https://github.com/harlamovads/RILEC}} 
(2)~We design a framework for L1 interference data augmentation, covering the generation of PPO-optimized GPT models, (\S\ref{sec:ppo_generation}), pattern-based rule generation (\S\ref{sec:gector_generation}), and prompt-based controlled generation (\S\ref{sec:llm_generation}).
(3)~Within this framework, we demonstrate and analyze capabilities and limitations of LLMs in erroneous sentences generation for data augmentation (\S\ref{sec:ppo_generation},\ref{sec:llm_generation}).  
(4)~We conduct a preliminary evaluation to assess the effect of each augmentation method on real data annotation (\S\ref{sec:results}) and release the best-performing model for future uptake.

\section{Related Work}

\paragraph{L1-Interference Error Detection}

L1 interference is the influence of the speaker's native language on their L2 production, either via direct transfer, or structural and semantic shifts \cite{weinreich1979languages,harris1995historical}. 
It is often shaped by linguistic similarity, leading to substitution or code-switching \cite{weinreich1979languages,kuvcuk2023error,dogruoz-etal-2021-survey}.
Recent works show that L1 features, such as identity, language family, and n-grams are useful in tasks like natural language inference (NLI) and grammatical error detection \cite{chang2008automatic,hermet-desilets-2009-using,tetreault-etal-2013-report,malmasi-etal-2017-report,rozovskaya2017adapting}. For instance, \citet{kochmar-shutova-2016-cross} demonstrate that L1 semantic information improves lexical error detection in verb-noun phrases. \citet{zomer-frankenberg-garcia-2021-beyond-grammatical} propose an L1-aware encoder-decoder model, outperforming GECToR and highlighting the need to integrate L1 features into developed grammatical error correction (GEC) systems.

\paragraph{Data Augmentation for GEC and GED}

The performance of GEC and GED models depends heavily on the size and quality of training data \cite{lichtarge-etal-2019-corpora, lichtarge2020data, nagata-etal-2022-exploring}. 
Synthetic augmented data is often used to scale small training corpora and to mitigate error type imbalance \cite{stahlberg-kumar-2021-synthetic, wang-etal-2024-improving-grammatical, lichtarge2020data}. Augmentation techniques are largely shared across GEC and GED, with tag-then-rewrite strategies proving particularly effective \cite{omelianchuk2020gector}. These techniques are commonly categorized into rule-based methods (noise injection and pattern matching) and model-based approaches (conditional generation and translation)\cite{kiyono2020massive, fang-etal-2023-transgec, ye-etal-2023-mixedit, stahlberg-kumar-2021-synthetic}.
\citet{wang-etal-2024-improving-grammatical} further introduce a contextual augmentation method that uses a model to regenerate context around error patterns extracted from existing parallel corpora. 
\citet{ye-etal-2023-mixedit} propose a mixed \textsc{MixEdit} approach, which regulates the similarity and diversity of generated data for augmentation quality improvement.

To the best of our knowledge, existing work has not addressed L1-specific data augmentation for the L1-GED task.
To bridge this gap, we build on prior research in data augmentation for GED, employing both rule-based and LLM-based methods to generate L1-influenced data for error detection.
We report a comparison of English learner and GEC corpora in \autoref{tab:rilec_corpora_comparison}.

\section{Corpus of L1-Interference Errors}\label{sec:l1_interference_errors}

Our corpus builds on the Russian Error-Annotated Learner of English Corpus (\textsc{REALEC}; \citetlanguageresource{vinogradova2022review}), which contains essays written by native Russian speakers in IELTS-like writing tasks, including descriptions of graphical data and opinion pieces, each under 300 words.
The corpus is manually annotated for errors, specifically for the presence of interference errors.\footnote{A detailed description of the annotation scheme and associated tags can be found in the official manual released with the corpus: \href{https://realec.org}{https://realec.org}.}
Each error is represented by a span supplied with a corresponding correction. 
The subset already annotated for the presence of interference errors (henceforth REALEC-L1) is used for baselines and augmentation.

\subsection{L1-Interference Annotation Scheme}
We use an annotation scheme for the L1-interference error tagging system derived from the framework of language interference analysis proposed by \citet{weinreich1979languages}. We describe the five-tag system in detail below.

\paragraph{Copying Expression} This category includes word-for-word translations of L1 expressions and collocations.
For instance, in (\ref{ex:CopyingExp1_unacc}), the Russian expression (\textit{kazhdovo iz nas}) is used as a substitute for an English linking phrase and is translated directly.
\begin{exe}
\ex \begin{xlist}
    \ex\label{ex:CopyingExp1_acc} A big bath was prepared for everyone.
    \ex\label{ex:CopyingExp1_unacc} * A big bath was prepared for \textbf{every of us}. (Bolshaya vanna byla prigotovlena dla \textbf{kazhdovo iz nas}.)
    \end{xlist}
\end{exe}

\paragraph{Synonyms} This category includes errors where the author intends to use a Russian word with multiple meanings corresponding to different English lexemes but selects the incorrect English word out of these corresponding words. 
For example, in (\ref{ex:Synonyms1_unacc}), the student mistakenly uses \textit{overcome} instead of \textit{cover}. 
Both of these English words correspond to the same Russian lexeme as provided in the translation \textit{preodolet'}. 

\begin{exe}
\ex \label{ex:Synonyms1_unacc} * The distance can be \textbf{overcame} by the train. (Distantsiya mozhet byt' \textbf{preodolena} poezdom.)
\end{exe}

(\ref{ex:Synonyms2_acc}) and (\ref{ex:Synonyms3_acc}) provide examples of contexts that use the same word \textit{preodolet'} in Russian, but correspond to two different words: \textit{overcome} and \textit{cover} in English.
We categorize this error as Synonyms, because the synonymous English words are confused due to the influence of a Russian word that encompasses both meanings.

\begin{exe}
\ex\label{ex:Synonyms2_acc} They covered the necessary distance in one day. (Oni preodoleli nuzhnoe rasstoyanie za odin den'.)
\ex\label{ex:Synonyms3_acc} They overcame all the challenges and emerged victorious.  (Oni preodoleli vse ispytanija i vyshli pobeditelami.)
\end{exe}

\paragraph{Tense Semantics} This type of error occurs when an English tense is used incorrectly due to the corresponding grammatical form in Russian.
In (\ref{ex:TenSem1_unacc}), we see that the present tense is used to describe graphical data relating to the past. While English requires the verb tense to align with the time frame of the event in such cases, in Russian, it is acceptable to use both past and present verb tenses for both historical present and graphical data description. This influences the learners and makes the confusion of past and present tenses in such contexts more frequent. Other errors of this type pertain to usage of \textit{will} in conditionals, etc.

\begin{exe}
\ex \begin{xlist}
    \ex\label{ex:TenSem1_acc} In 1999 the share decreased.
    \ex\label{ex:TenSem1_unacc} * In 1999 the share \textbf{decreases}. (V 1985 jeta dola \textbf{snizhaetsa'}.)
    \end{xlist}
\end{exe}

Note that we do not address issues arising from English tenses not present in Russian (e.g., Present Perfect vs. Past Simple), as these involve the absence of features in L1, unlike the presence of Russian tense features causing problems here.

\paragraph{Transliteration} This well-known type of error involves using Russian words written with the English alphabet in an English text. An example of such an error is shown in (\ref{ex:Transl1_unacc}), where \textit{cassa} is used instead of \textit{cashier}.

\begin{exe}
 \ex\label{ex:Transl1_unacc} * And often {a lot of} money {comes to}  \textbf{the cassa}. (Chasto mnogo deneg postupaet v \textbf{kassu}.)
\end{exe}

\paragraph{Word Form Transmission} This type of interference error involves transferring a grammatical category from Russian to English. In (\ref{ex:WFT1_unacc}), the form \textit{billions} is used erroneously most likely due to the equivalent Russian collocation which requires the plural form, with the plural marking carrying over into the English text.

\begin{exe}
 \ex\label{ex:WFT1_unacc} * The primary cost was {\$5 \textbf{billions}}. (Osnovnaya stoimost' sostavljala {5 \textbf{milliardov} dollarov}.)
\end{exe}

\begin{figure}[!ht]
    \centering
    \includegraphics[width=0.75\linewidth]{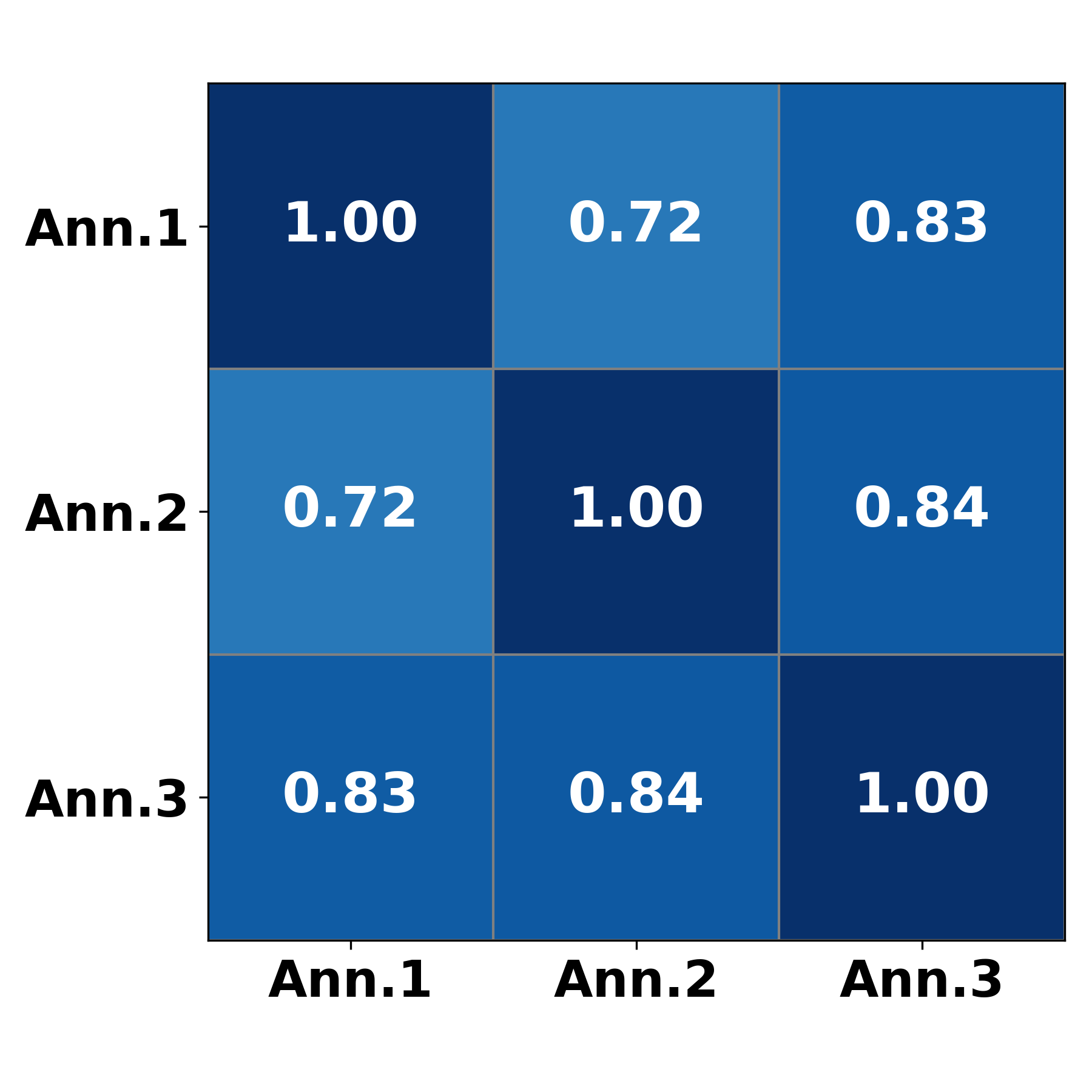}
    \caption{Pairwise inter-annotator agreement (Cohen's kappa) for the \textsc{REALEC-L1} data annotation.}
    \label{fig:iaa_agreement}
\end{figure}

The \textsc{REALEC-L1} subset contains 6,086 sentences, each annotated as exhibiting at least one interference error in the main \textsc{REALEC} corpus. To assess annotation reliability, we conduct an additional round on 500 sentences from \textsc{REALEC-L1}, carried out by four Russian-speaking linguists with C1–C2 proficiency in English. All annotators are thoroughly familiarized with the guidelines. 
We first conduct a preliminary calibration on 100 sentences to ensure a consistent understanding of the criteria. 
The annotations are then conducted independently, without communication between annotators.
Upon completion, we compute pairwise inter-annotator agreement (IAA) using Cohen’s kappa~\citep{doi:10.1177/001316446002000104}. The results, shown in \autoref{fig:iaa_agreement}, range from 0.72 to 0.84, indicating a high level of inter-annotator consistency.

The full \textsc{RILEC} dataset includes this subset and is further expanded using data augmentation techniques. 
Detailed corpus statistics by error type are presented in \S\ref{sec:corpus_overview} with generated examples in \autoref{tab:ungrammatical_examples}. 
We further describe how the synthetic data were generated and assess their contribution to classification performance.
The numbers of sentences in the train and test splits from \textsc{REALEC-L1} are reported in \autoref{tab:data_splits}.

\begin{table}[!ht]
\centering
\begin{tabular}{lcc}
\toprule
\textbf{Dataset} & \textbf{Train} & \textbf{Test} \\
\midrule
\textsc{REALEC-L1} & 3,882 & 2,204 \\
\midrule
\textsc{GPT-based (PPO)} & 4,583 & 1,965 \\
\textsc{LLM-based} & 556 & 239 \\
\textsc{Rule-based (GECToR)} & 4,131 & 1,771 \\
\midrule
\textbf{\textsc{RILEC}} & 12,652 & 6,179 \\
\bottomrule
\end{tabular}
\caption{L1-error dataset splits by train and test size. The GPT-, LLM-, and rule-based datasets represent synthetic augmented subsets.}
\label{tab:data_splits}
\end{table}

\section{Data Generation}\label{sec:data_generation}

We use three approaches for data augmentation:  
(1) small-scale LLMs optimized with Proximal Policy Optimization (PPO) to generate sentences similar to annotated examples;  
(2) a rule-based algorithm that introduces controlled errors by replacing correct words with incorrect ones;
(3) LLM prompting to generate erroneous sentences by modifying and replicating annotated interference errors.
We detail each method in the following subsections.

\subsection{PPO-based Generation}
\label{sec:ppo_generation}

In this section, we describe error generation with language models optimized using PPO \citep{schulman2017proximal}.

\paragraph{Model Set Up for PPO}
\label{sec:gpt2distil}
We experiment with two autoregressive models: GPT2\footnote{\href{https://huggingface.co/gpt2}{hf.co/gpt2}} and DistilGPT2\footnote{\href{https://huggingface.co/distilgpt2}{hf.co/distilgpt2}}. 
We first perform model fine tuning to adapt the models to in-domain generation using all sentences from \textsc{REALEC}. 
We observe that GPT2 is less affected by this step, as its next-word prediction loss decreases less than that of DistilGPT2, which performs better. 
We therefore continue with the latter model.

\paragraph{Reward Models}
To enable controlled generation of the five error types, we train a separate binary classifier for each error type, as described in \S\ref{sec:l1_interference_errors}. 
Each classifier is based on the RoBERTa-base model\footnote{\href{https://huggingface.co/FacebookAI/xlm-roberta-base}{hf.co/xlm-roberta}} fine tuned on the train subset of \textsc{REALEC-L1}. 
For each class, the dataset is divided into positive examples (containing the target error type) and negative examples (not containing it), with 80\% used for training and 20\% for evaluation. 
We fine tune each classifier for five epochs with a batch size of 16, a weight decay of 0.01, and a learning rate of 2$\times$10\textsuperscript{-5}.

\paragraph{Reinforcement Optimization with PPO}
Next, we apply PPO to optimize the fine tuned DistilGPT2 for controlled error generation.
Training is performed separately for each of the five error types, resulting in five optimized models. 
During training, the model generates sentences that are evaluated by the corresponding error specific classifier. 
The reward function assigns a positive score if the generated continuation contains the target error and a negative score otherwise. 
PPO is performed for five epochs with an 80/20 data split on the entire \textsc{REALEC} dataset containing L1 annotated sentences, using a batch size of 32 and a learning rate of 2$\times$10\textsuperscript{-5}. 
We find that DistilGPT2 classifier based reward scores increase during training, indicating successful alignment with the intended error generation.
We further analyze data generated by the five models trained for each L1 tag.
Sample continuations generated from short input segments are presented in \autoref{tab:ungrammatical_examples}.

\paragraph{Evaluation and Expert Feedback}
We manually analyze examples generated by the PPO optimized language models for each error type. 
To assess generation quality, we qualitatively annotate 50 randomly selected outputs per error type, marking the presence of the respective L1 influenced error.
Based on these annotations, we find that the generated samples correctly represent the target errors in all cases except for two categories. 
For \textit{Tense Semantics} and \textit{Transliteration}, the trained models fail to reliably generate errors. 
The former model does not produce suitable context for the erroneous verb form, while the latter generates random character sequences resembling transliteration errors (see Example~\ref{ex:Transl_fail}).

\begin{exe}
 \ex\label{ex:Transl_fail} Also, in the Middle East wich beaute a dramatost of hi teck and the femenest ode to alhtejut's exalt is so difunct.
\end{exe}

As a result, we exclude PPO-based generations for these two categories from the final dataset. 
For these categories, we only apply a rule-based generation strategy (see \S\ref{sec:rulebased}). 
To ensure natural sentence openings, we extract the most frequent sentence-initial words from the full \textsc{REALEC} corpus and use them as weighted prompts during generation.
We select prompts for generation from these words using random sampling weighted by their frequency in \textsc{REALEC}. 
This ensures that the initial word distribution closely matches the original dataset, avoiding over-representation of specific words that could bias later training.
In total, we obtain 6,547 examples by generating continuations for the three error tags based on query conditions, after filtering out sentences shorter than five tokens.

\begin{table*}[t]
\centering
\small
\resizebox{0.95\textwidth}{!}{
\begin{tabular}{lp{13cm}}
\toprule
\textbf{Error Type} & \textbf{Generation Examples} \\
\midrule

\error{copexp}{CopExp} & 
Overall all these places are \textbf{in our age}. (PPO) 

This is extremely unlikely for everyone and every scientist and medical worker in \textbf{influence on} world. (GECToR)

Males in \textbf{66-75 ages} made a slight jump compared to 56-65 years old males. (LLM)  \\[0.5em]

\error{synonyms}{Synonyms} & 
If such a person should enter this prison, they should be \textbf{given} what they had already done. (PPO)

In the one hand, people should make their living, and they should \textbf{take} the necessary education. (PPO)

This number is 1.5 times higher than the previous ones during the \textbf{departed} period. (GECToR)
\\[0.5em]
\error{wft}{WFT} & 
Also, the global total increase has fallen, and nearly 2 billion people, making it number \textbf{1 billion billions.} (PPO)

This will be \textbf{girls'} longest run we will see in nearly every post on this web. (GECToR) \\[0.5em]

\error{tensem}{TenSem} & 
It can be seen that in Africa the rate of unemployment \textbf{increases} at the same rate as in 2000, and then in Africa it \textbf{remains} at a steady rate. (Rule-Based)

There \textbf{is} no changes in this part of the population in the period in which 2000 was presented. (GECToR) 

The proportion of old individuals aged 66 and over \textbf{decrease} in periods 1950-1995 from 6\% to 4\% and start growing after that. (LLM) 
\\[0.5em]
\error{translit}{Transliteration} & 
The percent of \textbf{investitsment} came in Russia, where in 2015 it was 5 billion Euros and in 2018 it is 8 billion Euros. (Rule-Based) 

She is receiving lots of \textbf{funat} emails. (LLM)

This can be deduced from the fact that the highest rate of unemployment among South Asia \textbf{fabricks} Latin America in 2014 and 2015 was observed. (GECToR)
\\[0.5em]
\bottomrule
\end{tabular}}
\caption{Examples of ungrammatical model generations from \textsc{RILEC}, grouped by L1 error type. Data augmentation methods are indicated in brackets: PPO (Proximal Policy Optimization with DistilGPT2), GECToR (dictionary-based), Rule-based (tense replacement or transliteration), and LLM (prompt-based generation). \error{copexp}{CopExp} = Copying Expression; \error{wft}{WFT} = Word Form Transmission; \error{tensem}{TenSem} = Tense Semantics.}
\label{tab:ungrammatical_examples}
\end{table*}

\paragraph{Rule-based Error Injection}
\label{sec:rulebased}
We use a rule-based method to inject errors related to \textit{Tense Semantics} and \textit{Transliteration}.  
First, we use the fine tuned GPT2 (see \S\ref{sec:gpt2distil}) to obtain domain specific but mostly grammatically correct sentences (using the same probability based prompting strategy). 

For the \textit{Tense Semantics} tag, we filter sentences containing a year and modify the verb in the corresponding clause to the Present Simple form using the SpaCy package \citep{Honnibal_spaCy_Industrial-strength_Natural_2020}. 
As the generated years are typically associated with past or future events, this verb change results in an erroneous interpretation.
We generate this type of error because it reflects one of the most frequent error patterns found in the non synthetic \textsc{REALEC-L1} data, where past and future events are incorrectly described using the present tense when interpreting graphical data \citeplanguageresource{vinogradova2022review}.

For the \textit{Transliteration} tag, we replace randomly selected nouns with their transliterated versions obtained using the Google Translate API (accessed in July 2024).\footnote{\href{https://cloud.google.com/translate}{cloud.google.com/translate}}

Examples of both error types are presented in \autoref{tab:ungrammatical_examples}.
In total, we generate 1,748 augmented sentences for \textit{Tense Semantics} and 895 for \textit{Transliteration} using the described rule-based strategy.

\subsection{Error Generation with GECToR}\label{sec:gector_generation}

We adopt a rule-based approach following the one used for data augmentation for \textsc{GECToR} \cite{omelianchuk2020gector} training. Error generation used by \textsc{GECToR} authors focuses on introducing errors of three types: missing words, redundant words, and words that need to be replaced. We focus solely on implementing the replacement errors, as they are the most relevant for interference errors.

\paragraph{Data Preparation}
First, we collect possible corrections for the erroneous spans. We use the full \textsc{REALEC-L1} with native span corrections. To extend possible corrections, we utilize the XLM-RoBERTa model.\footnote{\href{https://huggingface.co/FacebookAI/xlm-roberta-base}{hf.co/xlm-roberta}} We mask the error spans in the base sentences and treat the suggested filled masks as corrections for the respective sentences. After that, we pair the corrections and the error spans to create a dictionary of possible errors: each correction is mapped to a list of tokens that could erroneously replace it. For each error type, we create its own corrections dictionary.

\paragraph{Error Injection}
We adjust \textsc{GECToR}'s official implementation so that, for every sentence, a single word is chosen and replaced in accordance with the dictionaries. After the error is introduced, the code marks it and specifies its type based on the collection that was sourced to introduce the error. We run the code on grammatically correct generations from the GPT2-based generator after fine-tuning on \textsc{REALEC-L1}, as described in \S\ref{sec:ppo_generation}, since it was found to produce thematically suitable and mostly grammatically correct sentences.
To classify errors into five types, we use a baseline classifier fine-tuned on \textsc{REALEC-L1}, using the same experimental settings as in \S\ref{sec:results}.
As a result, we obtain 5,900 augmented examples through rule-based error injection.
Examples of the generated data are provided in \autoref{tab:ungrammatical_examples}.


\subsection{Prompted Generation of Interference Errors}\label{sec:llm_generation}

To generate new examples for each interference tag, we prompt decoder-only LLMs with class details for each interference type with examples.
After generating data with the best performing model, we perform error-annotation using LLMs.

\paragraph{Data Generation}

We begin by selecting the best model for generation through annotation among LLMs designed for open-ended conversations. The three models we select for experiments are accessible via API\footnote{Accessed in July 2024.} include: Claude 2,\footnote{\href{https://www.anthropic.com/index/introducing-claude}{anthropic.com/introducing-claude}} GPT-3.5,\footnote{\url{https://chat.openai.com}} and Mistral~\cite{jiang2023mistral7b}.\footnote{\url{https://mistral.ai}} We prompt the models to generate new examples using the following input, concatenated with 10 randomly selected sentences from the corpus manually annotated for L1 interference class errors:

\begin{small}
\begin{flushleft}
Here are some sentences with L1-motivated mistakes. Find the mistakes in these sentences and generate new contexts with different meanings, while retaining the mistakes from the original sentences.
\end{flushleft}
\end{small}
For generation, we set the temperature to 1.0 and use the default repetition penalty for each model.
We paraphrase the prompt four times and repeat generation for each version.  
Overall, we obtain 40 erroneous sentences for each of the models. 

Next, we ask an expert specializing in error annotation to review the generated model outputs.  
The expert marks a generation as successful if it contains an error of the same type as in the source sentence and as unsuccessful if the present error differs from the one in the provided sentence.  
From the annotation results, we find that the Claude 2 model outperforms the rest, with 38 out of 40 successful generations.  
Mistral successfully generates examples in half of the cases, whereas GPT-3.5 does not follow the instructions and fails to generate erroneous sentences at all.

We select Claude 2 for data generation and generate 800 examples, each time using 10 newly selected samples from the source corpus.  
The generation process takes a few hours, with necessary pauses due to per-hour generation limits.  
We then proceed with the annotation of the generated examples.

\paragraph{Evaluation}

Given the generated examples, we repeat the procedure, but this time for annotating the examples.  
We select the model optimal for annotation with assistance of the expert.
First, we prompt the models with annotation instructions to identify and label errors of a given type.
For annotating examples, we follow the instructions from the source \textsc{REALEC-L1} dataset, originally designed for expert annotators and construct the following prompt:

\begin{small}
\begin{flushleft}
Here are sentences that contain mistakes.  
Some mistakes are caused by interference with the Russian language.  
Find and highlight such mistakes. Classify the mistakes according to the \textit{Instructions}.
The following sentences are provided as examples.
\end{flushleft}
\end{small}

We annotate 40 randomly selected sentences from the source data using this prompt for three models  
and ask the linguistic expert to review the annotated outputs.  
The expert marks their agreement with the error span and the type of L1 interference specified by the model,  
both of which we consider together as an accurate annotation.  
We find that the Mistral model outperforms the rest, with 40 out of 40 correct annotations.  
Meanwhile, Claude and GPT-3.5 correctly annotate fewer than 10 cases. We therefore select Mistral model for data annotation.  
Using this model, we annotate 800 examples generated by Claude-2 previously selected for data generation.

As a result, we obtain 794 annotated examples out of 800 previously generated erroneous sentences. Six sentences were excluded at this stage due to very closely resembling other LLM-generated sentences.

\begin{table*}[t]
\small
\centering
\begin{tabular}{lccccccc}
    \toprule
    \multirow{2}{*}{\textbf{Dataset}}& \multirow{2}{*}{\textbf{$F1_{train}$}} & \multicolumn{6}{c}{\textbf{\textbf{$F1_{test}$} Tag-specific}} \\
    \cmidrule(r){3-8}
    & & \textbf{CopExp} & \textbf{Syn} & \textbf{TenSem} & \textbf{Transl.} & \textbf{WFT} & \textbf{Avg.}\\
    \midrule
    Base  & 55.66 & 15.79 & 46.60 & 75.44 & 83.33 & 57.14 & 55.66 \\
    GPT-Based & \textbf{83.76} & 21.62 & \textbf{79.25} & \textbf{82.14} & 92.31 & \underline{83.72} & \underline{71.81} \\
    Rule-Based & 78.53 & 31.82 & 61.40 & 77.06 & 84.62 & 75.00 & 65.98 \\
    LLM-Based  & 44.78 & \textbf{41.03} & \underline{72.16} & 78.85 & \underline{92.31} & 71.43 & 71.16 \\
    \midrule
    \textbf{\textsc{RILEC}} & \underline{83.00} & \underline{33.33} & 69.39 & \underline{80.00} & \textbf{96.55} & \textbf{90.48} & \textbf{73.95}\\
    \bottomrule
\end{tabular}
\caption{L1-interference error classification results for RoBERTa models fine-tuned on augmented subsets: augmented with DistilGPT optimized with PPO (GPT-based; \S\ref{sec:ppo_generation}), with prompting LLMs (LLM-based; \S\ref{sec:llm_generation}), and obtained via rule-based generation (\S\ref{sec:gector_generation}). We report average F1-scores on the training dataset and tag-specific scores. CopExp = Copying Expression; WFT = Word Form Transmission; TenSem = Tense Semantics. The best score is in bold, and the second-best score is underlined.}
\label{tab:performance}
\end{table*}

\subsection{Corpus Overview}\label{sec:corpus_overview}
\begin{figure}[ht]
    \centering\includegraphics[width=0.48\textwidth]{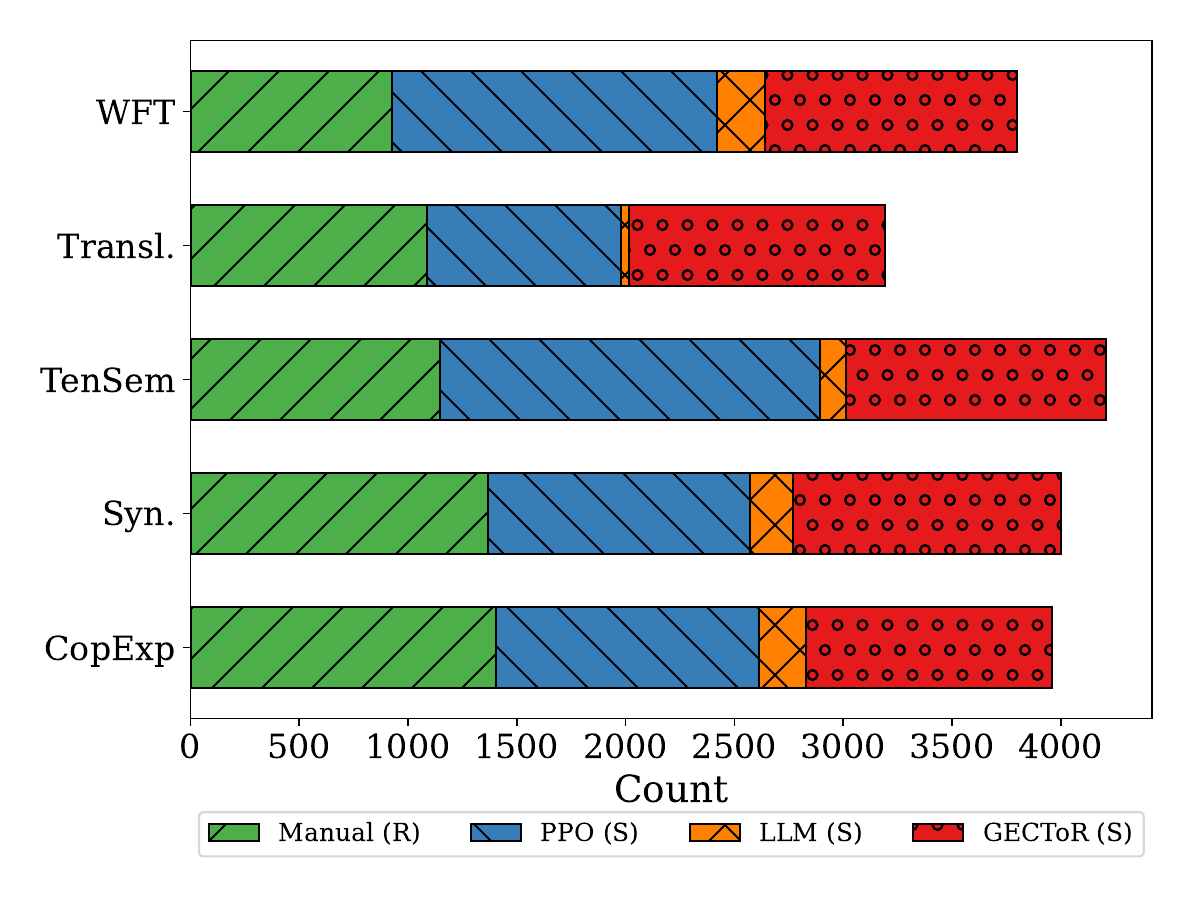}
\caption{Error distribution in \textsc{RILEC}, including synthetic (S) and real (R) data. CopExp = Copying Expression; WFT = Word Form Transmission; TenSem = Tense Semantics.}
    \label{fig:distribution_rilec}
\end{figure}

\autoref{fig:distribution_rilec} shows the distribution of errors by type and generation method in \textsc{RILEC}. With the described data augmentation, we obtain approximately 4,000 erroneous texts for each L1 interference type, resulting in a total of 18,830 sentences. 
This dataset is subsequently used to train an L1 error classification model. 
In addition, we evaluate the diversity of the synthetic data relative to the original data using Self-BLEU, MAUVE, and 3-gram novelty, as reported in \autoref{sec:rilec_diversity}.

\section{L1-interference Error Classification}\label{sec:results}
In this section, we report results of models trained with different augmented datasets and analyze the performance of the resulting L1 interference error detection models.

\subsection{Experimental Settings}
\label{sec:exp_settings}

We approach L1-interference error classification as a multi-span classification task and implement the solution using the \textsc{SpaCy} span classification pipeline.\footnote{\url{https://v2.spacy.io/usage}}  
We perform experiments using the RoBERTa-base model\footnote{\href{https://huggingface.co/FacebookAI/roberta-base}{hf.co/FacebookAI/roberta-base}}.  
Each model is fine tuned with a batch size of 128, a span classification threshold of 0.5, and trained for ten epochs.

\paragraph{Data}
For each augmentation method, we shuffle and split the data, using 80\% for training and 20\% for testing.  
The dataset split statistics are reported in \autoref{tab:data_splits}.  
To estimate the contribution of augmented data to \textsc{REALEC-L1} classification, we evaluate all models on the \textsc{REALEC-L1} test set.  
We also report the F1-score on the in-domain data used for fine tuning.  
As a baseline, we use the model trained on the \textsc{REALEC-L1} dataset described in \S\ref{sec:l1_interference_errors}.

\subsection{Results}\label{sec:results_results}

\autoref{tab:performance} reports the error detection performance for each L1-interference category. The classifier trained on \textsc{RILEC}, which includes all augmented data, substantially outperforms the baseline and models fine-tuned on individual augmentation subsets. The highest F1-scores are observed for \textit{Transliteration} and \textit{Word Form Transmission}, both exceeding 90\%, with an overall average of approximately 74\%, compared to 55.66\% for the baseline.

\paragraph{Effect of Augmented Data}

Models fine-tuned on data generated through model-based augmentation methods outperform those trained on rule-based data, despite the larger size of the latter. 
The PPO-optimized \textsc{DistilGPT} model achieves the second-highest average F1-score (71.81\%), followed closely by the LLM-prompted model, despite using significantly fewer training examples. The LLM-based model also exhibits better generalization, as indicated by lower training and higher test performance.

\paragraph{Class-Specific Performance}

The model fine-tuned on the full \textsc{RILEC} dataset achieves high F1-scores (above 80\%) on the \textit{Tense Semantics}, \textit{Transliteration}, and \textit{Word Form Transmission} classes. In contrast, performance is lower on \textit{Copying Expression} (33.33\%) and \textit{Synonyms} (69.39\%), which require more complex handling of L1-specific lexical and collocational interference. 
For these categories, the best-performing models are those fine-tuned on LLM-prompted data and PPO-optimized \textsc{DistilGPT}, suggesting they are more effective at modeling such interference patterns.

\paragraph{Manual Analysis}
We conduct a manual evaluation with a linguist on two datasets: 100 test sentences ($Test_{100}$) and 70 randomly sampled sentences from the \textsc{REALEC} learner corpus (\textsc{REALEC}$_{70}$). 
This allows us to compare model performance on benchmark and real learner data. 
For $Test_{100}$, we report accuracy based on both gold annotations and expert-approved alternatives, including \textit{Distinct TPs}—correct predictions unique to each model. 
For \textsc{REALEC}$_{70}$, we report true positives and TP–FP (net correct) counts. 
\autoref{tab:results_manual} shows that all models trained on augmented data outperform the baseline, with the highest scores (73–74\%) achieved by models trained on \textsc{RILEC} and PPO-optimized GPT outputs.

\begin{table}[h]
\footnotesize
\centering
\begin{tabular}{lcccc}
    \toprule
    \multirow{2}{*}{\textbf{Model}} & \multicolumn{2}{c}{\textbf{Test$\mathbf{_{100}}$}} & \multicolumn{2}{c}{\textbf{\textsc{REALEC}$_{70}$}} \\
    \cmidrule(lr){2-3} \cmidrule(lr){4-5}
    & \textbf{Acc.} & \textbf{Dist.} & \textbf{TP} & \textbf{TP–FP} \\
    \midrule
    Base           & 0.66 & 0  & 6  & 4 \\
    GPT-based      & 0.74 & 14 & 15 & 7 \\
    Rule-based     & 0.74 & 10 & 17 & 13 \\
    LLM-based      & 0.68 & 8  & 8  & 5 \\
    \midrule
    \textbf{\textsc{RILEC}} & 0.73 & 10 & 19 & 10 \\
    \bottomrule
\end{tabular}
\caption{Manual evaluation. Test$_{100}$: Acc.~= accuracy, Dist.~= distinct TPs. \textsc{REALEC}$_{70}$: TP = true positives, TP–FP = net correct predictions.}
\label{tab:results_manual}
\end{table}

The linguist notes that the rule-based model performs particularly well on tags such as \textit{Synonyms} (\ref{ex:synonym_rule}) and \textit{Copying Expression} (\ref{ex:copying_rule}), which show the lowest overall scores in \autoref{tab:results_manual}.

\begin{exe}
\ex \label{ex:synonym_rule}
...the population of Sweden \textbf{outlived a considerable growth} (\textbf{perezhila znachitel’nyi rost})
\ex \label{ex:copying_rule}
... we can not only \textbf{safe} our time, but also it allows us to \textbf{achieve to} our destination.
(\textbf{sokhranit'}, \textbf{dostignut'})
\end{exe}

The model fine-tuned on optimized DistilGPT data achieves the highest number of distinct correct predictions (14), capturing valid errors missed by other models.

On \textsc{REALEC}$_{70}$, model performance follows a similar trend: \textsc{RILEC}-trained model achieves the highest true positive count (19), followed by the pattern-generated model (17), both outperforming the baseline (6). The same models also lead in terms of the TP–FP balance. The model fine-tuned on prompting data exceeds the baseline by a smaller margin, possibly due to sensitivity to domain mismatch.

Overall, our results demonstrate that the model fine-tuned on \textsc{RILEC}, which includes all augmented data, outperforms the baseline by a large margin in classification scores averaged across all L1 interference error classes. It achieves particularly high F1-scores on Word Form Transmission, Transliteration, and Tense Semantics errors.  Manual analysis further confirms that the RoBERTa model fine-tuned on \textsc{RILEC} successfully detects errors in new learner texts, validating the effectiveness of the designed span annotation model in practice.

\section{Conclusion}

In this work, we introduce \textsc{RILEC}, a large-scale corpus for detecting L1 interference errors in English essays written by Russian-speaking learners. 
The corpus combines manually annotated data from \textsc{REALEC} with synthetically generated data created using rule-based and language model-based augmentation methods. 

To assess the impact of data augmentation, we trained models on different corpus subsets and observe consistent performance gains, especially when using PPO optimization. 
Manual evaluation confirmed that augmented data better capture the semantics, style, and grammar of real learner language. 
Models trained on \textsc{RILEC} performed best on word-level interference types such as transliteration, tense semantics, and grammatical form mismatches, while showing lower accuracy on collocational and semantic categories like synonym substitutions and copying expressions.

In future work, we plan to extend \textsc{RILEC} with essays written by learners from diverse L1 backgrounds, using resources such as ICNALE and native language identification datasets~\cite{hermet-desilets-2009-using,ishikawa2018icnale}. 
We also plan to extend the evaluation to generative models. 
Preliminary experiments with GPT-5 and Claude indicate that these models exhibit low precision in L1 error detection on the \textsc{RILEC} data.


%

\section*{Limitations}

This study focuses on the detection of L1-motivated errors using a fixed five-tag annotation scheme. While this allows for systematic evaluation, it limits the granularity of linguistic phenomena captured. Future work could extend this by analyzing part-of-speech distributions and syntactic dependency relations, which may help uncover deeper patterns of L1 interference and improve classification performance.

We also find that GPT2, even when fine-tuned and trained with PPO, fails to reliably generate certain error types, particularly Tense Semantics and Transliteration errors. 
This suggests that GPT2 resists producing ungrammatical or unnatural constructions in these categories. 
To address this, we implement rule-based augmentation strategies. Although effective, this introduces an external dependency and limits possible generation quality and diversity. 
Future work could explore alternative model compression techniques, such as quantization or pruning, to better adapt small models like GPT2 for controlled error generation.

Finally, our dataset is restricted to IELTS-style English essays written by Russian-speaking learners. 
While the proposed framework is applicable to other L1 groups and genres, we do not evaluate cross-linguistic generalizability in this work. 

\section*{Ethical Considerations}

This study uses anonymized, publicly available data from the \textsc{REALEC} corpus, in accordance with its educational and research license. No personal or sensitive information is included. 
Our released models support language learning by identifying L1-specific error patterns, not for grading or assessment. We emphasize that automated error detection should be used with human oversight. 
Furthermore, model-based data augmentation methods could be misused, for instance, to generate fake learner data, simulate interference patterns, or fabricate responses on learning platforms. 
While such risks are limited, we highlight responsible use and restrict our models to educational and research purposes. Future work may expand the dataset to cover more L1 backgrounds to improve generalizability and support learners from diverse linguistic backgrounds.

\section{Bibliographical References}\label{sec:reference}

\bibliographystyle{lrec2026-natbib}
\bibliography{lrec2026-example}

\begin{thebibliography}{9}
\expandafter\ifx\csname natexlab\endcsname\relax\def\natexlab#1{#1}\fi

\bibitem[{Blanchard et~al.(2013)Blanchard, Tetreault, Higgins, Cahill, and Chodorow}]{blanchard2013toefl11}
Blanchard, Daniel and Tetreault, Joel and Higgins, Derrick and Cahill, Aoife and Chodorow, Martin. 2013.
\newblock \emph{TOEFL11: A corpus of non-native English}.
\newblock Wiley Online Library.

\bibitem[{Bryant et~al.(2019)Bryant, Felice, Andersen, and Briscoe}]{bryant-etal-2019-bea}
Bryant, Christopher and Felice, Mariano and Andersen, {\O}istein E. and Briscoe, Ted. 2019.
\newblock \href {https://doi.org/10.18653/v1/W19-4406} {\emph{The {BEA}-2019 Shared Task on Grammatical Error Correction}}.
\newblock Association for Computational Linguistics.

\bibitem[{Granger et~al.(2020)Granger, Dupont, Meunier, Naets, and Paquot}]{granger2020iclev3}
Granger, Sylviane and Dupont, Maïté and Meunier, Fanny and Naets, Hubert and Paquot, Magali. 2020.
\newblock \href {https://dial.uclouvain.be/pr/boreal/object/boreal:229877} {\emph{The International Corpus of Learner English. Version 3}}.
\newblock Presses universitaires de Louvain.

\bibitem[{Ishikawa(2023)}]{ishikawa2023icnale}
Ishikawa, Shin'ichiro. 2023.
\newblock \emph{The ICNALE guide: An introduction to a learner corpus study on Asian learners’ L2 English}.
\newblock Routledge.

\bibitem[{Nagata et~al.(2011)Nagata, Whittaker, and Sheinman}]{nagata2011creating}
Nagata, Ryo and Whittaker, Edward and Sheinman, Vera. 2011.
\newblock \emph{Creating a manually error-tagged and shallow-parsed learner corpus}.

\bibitem[{Napoles et~al.(2017)Napoles, Sakaguchi, and Tetreault}]{napoles-etal-2017-jfleg}
Napoles, Courtney and Sakaguchi, Keisuke and Tetreault, Joel. 2017.
\newblock \href {https://aclanthology.org/E17-2037/} {\emph{{JFLEG}: A Fluency Corpus and Benchmark for Grammatical Error Correction}}.
\newblock Association for Computational Linguistics.

\bibitem[{Ng et~al.(2014)Ng, Wu, Briscoe, Hadiwinoto, Susanto, and Bryant}]{ng-etal-2014-conll}
Ng, Hwee Tou and Wu, Siew Mei and Briscoe, Ted and Hadiwinoto, Christian and Susanto, Raymond Hendy and Bryant, Christopher. 2014.
\newblock \href {https://doi.org/10.3115/v1/W14-1701} {\emph{The {C}o{NLL}-2014 Shared Task on Grammatical Error Correction}}.
\newblock Association for Computational Linguistics.

\bibitem[{Vinogradova and Lyashevskaya(2022)}]{vinogradova2022review}
Vinogradova, Olga and Lyashevskaya, Olga. 2022.
\newblock \emph{Review of practices of collecting and annotating texts in the learner corpus REALEC}.
\newblock Springer.

\bibitem[{Yannakoudakis et~al.(2018)Yannakoudakis, Andersen, Geranpayeh, Briscoe, and Nicholls}]{yannakoudakis2018developing}
Yannakoudakis, Helen and Andersen, {\O}istein E and Geranpayeh, Ardeshir and Briscoe, Ted and Nicholls, Diane. 2018.
\newblock \emph{Developing an automated writing placement system for ESL learners}.
\newblock Taylor \& Francis.

\end{thebibliography}

\section{Language Resource References}
\label{lr:ref}
\bibliographystylelanguageresource{lrec2026-natbib}
\bibliographylanguageresource{languageresource}

\newpage
\appendix
\section{Experimental Settings}\label{sec:datageneration_appendix}

All training and fine-tuning experiments were conducted on a single NVIDIA A100 GPU with 80\,GB of GPU memory.

\paragraph{Model Fine-tuning}
We fine-tune GPT2 and DistilGPT2 using an 80/20 train–test split for 3 epochs with a batch size of 32, a learning rate of \(2 \times 10^{-5}\), and weight decay of 0.01.

\paragraph{PPO Optimization and Reward Models}
PPO optimization and reward model training are performed for 5 epochs with a learning rate of \(1.41 \times 10^{-5}\) and a batch size of 16. Sampling uses a temperature of 0.7 with top-\(k=0\) and top-\(p=1.0\).

\section{Lexical Diversity}\label{sec:rilec_diversity}

To ensure that the augmentation pipelines enrich the source REALEC-L1 data distribution, we evaluate the diversity of data generated by the rule-based, LLM-based, and GPT-based pipelines.
We evaluate Self-BLEU \citep{zhang2017adversarial}, MAUVE \citep{pillutla2021mauve}, and 3-gram novelty on the downsampled sentences relative to the REALEC-L1 test set of 2,204 sentences. 
\autoref{tab:diversity} reports the evaluation results for the generated data.
We find that the rule-based pipeline achieves the highest 3-gram novelty ($0.92$), indicating strong lexical novelty relative to the source distribution. The LLM-based pipeline yields the highest MAUVE score ($0.97$), suggesting the closest distributional alignment with REALEC-L1, while also exhibiting the lowest Self-BLEU ($12.24$), reflecting greater internal diversity. The GPT-based pipeline produces moderate scores across all metrics.
Overall, these results demonstrate that the proposed augmentation pipelines successfully generate diverse, non-redundant error patterns that enrich the original data distribution.

\begin{table}[h]
\centering
\scriptsize
\resizebox{\linewidth}{!}{
\begin{tabular}{lccc}
\toprule
\textbf{Pipeline} & \textbf{Self-BLEU $\downarrow$} & \textbf{MAUVE $\uparrow$} & \textbf{3-gram Novelty $\uparrow$} \\
\midrule
GPT-based & 15.25 & 0.47 & 0.70 \\
LLM-based & 12.24 & 0.97 & 0.20 \\
Rule-based & 12.75 & 0.36 & 0.92 \\
All Augmented & 14.03 & 0.48 & 0.92 \\
\midrule
REALEC-L1 & 16.37 & 1.00 & 0.00 \\
\bottomrule
\end{tabular}
}
\caption{Diversity and distributional similarity of augmented datasets relative to REALEC-L1. Lower Self-BLEU indicates greater diversity, while higher MAUVE and 3-gram Novelty indicate closer distributional alignment and higher n-gram novelty.}
\label{tab:diversity}
\end{table}

\end{document}